\documentclass[10pt]{article}
\usepackage{bofflab}

\usepackage{microtype}
\usepackage{parskip}

\usepackage{booktabs}
\usepackage{caption}
\usepackage{multirow}
\usepackage{wrapfig}
\usepackage{makecell}
\usepackage{adjustbox}
\usepackage{fullpage}
\usepackage{ragged2e}

\usepackage{placeins}                 
\usepackage{algorithm}
\usepackage{algorithmic}

\newtcolorbox{samplebox}[1]{
	enhanced,
	colback=LinkBlue!5!white,
	colframe=black!60,
	boxrule=0.6pt,
	arc=6pt,
	left=4pt, right=4pt, top=4pt, bottom=4pt,
	boxsep=3pt,
	before upper={\noindent\textbf{\small #1}\par\smallskip},
}

\usepackage{amsmath}
\usepackage{amssymb}
\usepackage{amsfonts}
\usepackage{bm}
\usepackage{mathtools}
\usepackage{amsthm}
\usepackage{cancel}
\usepackage{enumitem}

\DeclareMathOperator*{\argmax}{argmax}
\setlist[enumerate,1]{leftmargin=2em, itemsep=1em, parsep=0pt}
\setlist[itemize,1]{leftmargin=2em}

\usepackage[capitalize,noabbrev]{cleveref}
\usepackage{thm-restate}
\crefformat{section}{Section~#2#1#3}
\crefformat{subsection}{Section~#2#1#3}
\crefformat{subsubsection}{Section~#2#1#3}
\crefformat{equation}{(#2#1#3)}
\crefrangeformat{equation}{(#3#1#4) to (#5#2#6)}
\crefmultiformat{equation}{(#2#1#3)}{ and (#2#1#3)}{, (#2#1#3)}{ and (#2#1#3)}
\crefname{appendix}{Appendix}{Appendices}
\Crefname{appendix}{Appendix}{Appendices}
\crefformat{appendix}{Appendix~#2#1#3}
\AddToHook{cmd/appendix/before}{%
\crefalias{section}{appendix}%
\crefalias{subsection}{appendix}%
\crefalias{subsubsection}{appendix}
}
\crefname{table}{Table}{Tables}
\crefname{figure}{Figure}{Figures}
\crefname{lemma}{Lemma}{Lemmas}
\crefname{assumption}{Assumption}{Assumptions}

\definecolor{d62728}{RGB}{214,39,40}
\definecolor{1f77b4}{RGB}{31,119,180}
\definecolor{9467bd}{RGB}{148,103,189}
\definecolor{FigNY}{HTML}{1a1a3e}    
\definecolor{FigSD}{HTML}{b0a8d0}    
\usepackage{colortbl}
\definecolor{pinegreen}{rgb}{0.0, 0.47, 0.44}
\definecolor{cornellred}{rgb}{0.7, 0.11, 0.11}
\definecolor{cadmiumgreen}{rgb}{0.0, 0.42, 0.24}
\definecolor{spirodiscoball}{rgb}{0.06, 0.75, 0.99}
\definecolor{blizzardblue}{rgb}{0.73, 0.96, 0.99}
\definecolor{aliceblue}{rgb}{0.91, 0.94, 0.97}
\definecolor{darkblue}{RGB}{232, 224, 240} 
\definecolor{Red7}{rgb}{0.941, 0.243, 0.243}
\definecolor{Green7}{RGB}{55, 178, 77}

\makeatletter
\def\thm@space@setup{%
  \thm@preskip=0.8em plus 0.2em minus 0.2em
  \thm@postskip=0.8em plus 0.2em minus 0.2em
}
\makeatother

\theoremstyle{plain}
\newtheorem{theorem}{Theorem}[section]

\theoremstyle{definition}
\newtheorem{definition}[theorem]{Definition}

\makeatletter
\let\c@table\c@figure
\let\c@algorithm\c@figure
\makeatother

\usepackage{pifont}
\usepackage{array}
\usepackage{xurl}

\newcolumntype{C}[1]{>{\centering\arraybackslash}p{#1}}
\newcolumntype{L}[1]{>{\raggedright\arraybackslash}p{#1}}

\renewcommand{\Phi}{\boldsymbol{\varPhi}}      
\newcommand{\phivec}{\boldsymbol{\phi}}        
\newcommand{\phistar}{\boldsymbol{\phi}^*}     
\newcommand{\DeltaPhi}{\Delta\boldsymbol{\varPhi}}
\newcommand{\thetavec}{\boldsymbol{\theta}}    
\newcommand{\Reward}{R}                         
\newcommand{\Traj}{\tau}                        
\newcommand{\Trajfail}{\tau^-}                  
\newcommand{\Critic}{\mathcal{C}}              

\newcommand{\dimP}{\texttt{P}}      
\newcommand{\dimT}{\texttt{T}}      
\newcommand{\dimS}{\texttt{S}}      
\newcommand{\dimMid}{\texttt{Mid}}  
\newcommand{\dimM}{\texttt{M}}      

\newcommand{\phit}{\boldsymbol{\phi}_t}        
\newcommand{\thetat}{\boldsymbol{\theta}_t}    
\newcommand{\Dt}{\mathcal{D}_t}               
\newcommand{\AutoHarnessLoop}{\textsc{AutoHarnessLoop}}
\newcommand{\ModelAlignLoop}{\textsc{ModelAlignLoop}}

\title{Co-Harness: Co-Evolving Harnesses and Model Weights for LLM Agents}

\author[1,\dagger]{Zhengyu Chen}
\author[2,\dagger]{Teng Xiao}
\author[1]{Huaisheng Zhu}
\author[3]{Yige Yuan}
\author[1]{Luan Zhang}
\author[1]{Jingang Wang}

\affiliation[1]{Meituan}
\affiliation[2]{Allen Institute for AI}
\affiliation[3]{Independent}

\contribution[\dagger]{Equal contribution.}

\begin{document}
\begin{abstract}
Post-training agents for automated AI research requires optimizing not only model parameters, but also the runtime harness that shapes how research trajectories are generated, evaluated, and learned from. Existing pipelines typically train models under a fixed harness, including prompts, tools, skills, middleware, and memory, while leaving the data-generating process outside the optimization objective. This creates a mismatch between model updates and the static scaffolding that determines trajectory quality. We introduce Co-Harness, a framework that jointly optimizes the agent harness and model parameters during post-training. Co-Harness alternates between harness optimization and model optimization. An LLM-based HarnessCritic analyzes failed trajectories, identifies harness-level failure modes, and proposes validated local updates. The model is then fine-tuned on high-quality trajectories generated by the improved harness, distilling effective scaffolding into model parameters. A 200+ hour autonomous case study further shows that Co-Harness can recover from system crashes, improve inference efficiency, and discover ensemble strategies without human intervention. These results suggest that joint harness and model optimization is an effective way to improve  agents beyond fixed-harness post-training.

\end{abstract}
\maketitle

\section{Introduction}
\label{sec:introduction}

\begin{wrapfigure}{r}{0.6\linewidth}
  \centering
  \vspace{-1em}
  \includegraphics[width=\linewidth]{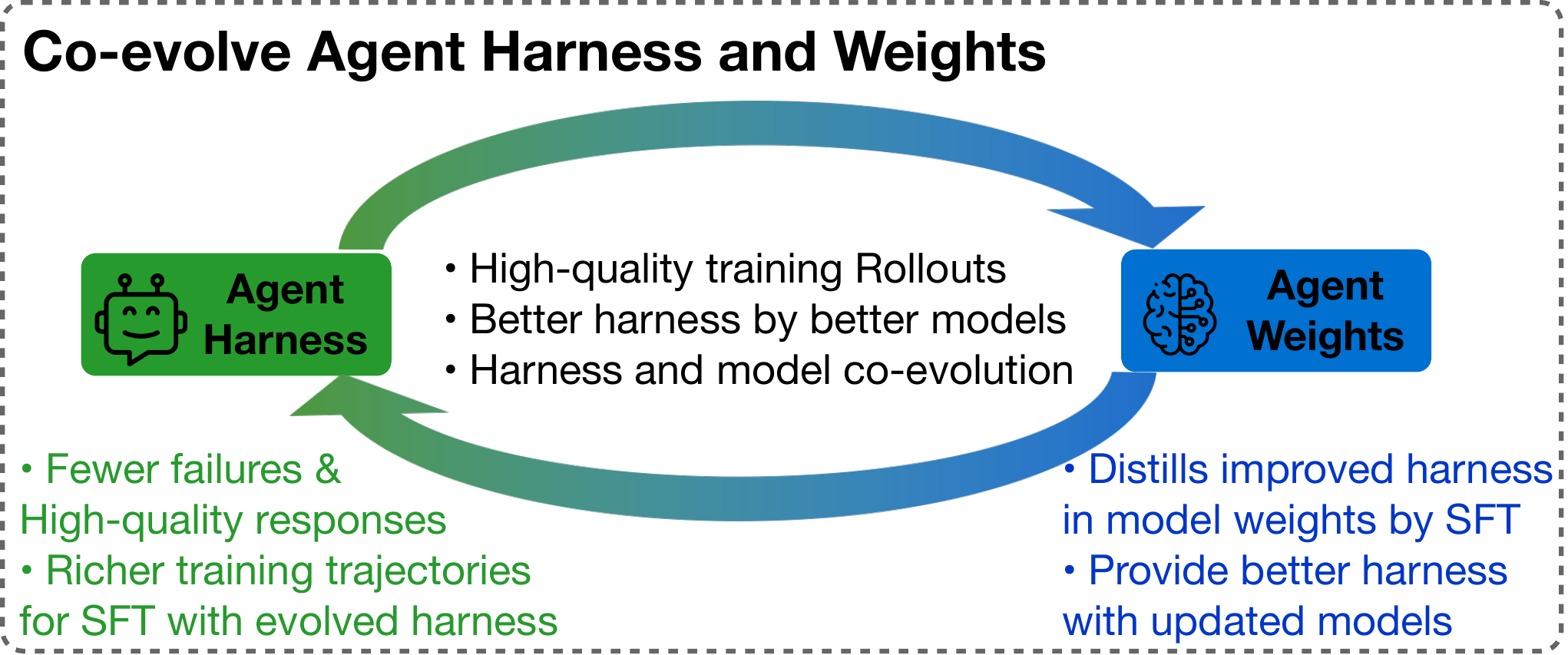}
  \caption{\textbf{Co-Harness dual-loop overview.}}
  \label{fig:co-harness-intro}
  \vspace{-1em}
\end{wrapfigure}

Post-training an agent is not only about training model weights, but also about designing the runtime system around the model. In practice, an agent consists of two coupled components: model parameters and the \emph{Harness} that elicits, executes, verifies, and records its behavior. The Harness includes prompts, tools, reusable skills, middleware, retry logic, context management, and memory. Current post-training pipelines treat these components asymmetrically. The model is updated through supervised fine-tuning, reinforcement learning, or preference optimization, while the Harness is usually hand-designed and fixed. This creates a mismatch: the Harness determines the trajectories used for training, yet it is not improved from failures observed during training. A bad tool schema, missing retry hook, short turn limit, or underspecified prompt can prevent useful trajectories from being generated. As models become more capable in software engineering and tool-use settings, the fixed Harness becomes an important bottleneck for automating AI R\&D \citep{amodei2024machines}.

Recent work suggests that agent performance can be improved by editing the system around the model, rather than the model itself.
One line of work exposes a narrow interface for such edits: prompts, demonstrations, or language-model programs are revised using execution feedback~\citep{khattab2023dspy,opsahl2024optimizing,yuksekgonul2025optimizing,agrawal2025gepa}, and in-context agent systems store past experience as textual artifacts that are reused at test time~\citep{zhao2024expel,zhang2025agentic}.
A newer line broadens this editable interface from a single prompt-like component to the full agent Harness.
Meta-Harness~\citep{leeMetaHarnessEndtoEndOptimization2026} optimizes harness code from source code, scores, and traces.
Agentic Harness Engineering~\citep{lin2026agentic} performs observability-driven updates to prompts, tools, middleware, skills, and memory.
AEVO~\citep{zhang2026harnessing} treats agentic evolution as an interactive process and edits the procedure or context that drives future search.
Together, these works make a clear case that the Harness is not just infrastructure, but an optimization variable.
However, the optimized Harness is still wrapped around a fixed model.
It improves how the current model runs at inference time, but it is not part of the post-training loop that produces trajectories, filters them, and updates the next model.
This motivates our question: \textit{can harness evolution be coupled with model post-training, so that the system that generates training data improves together with the model trained on that data?}

\begin{figure}[t]
  \centering
  \includegraphics[width=\linewidth]{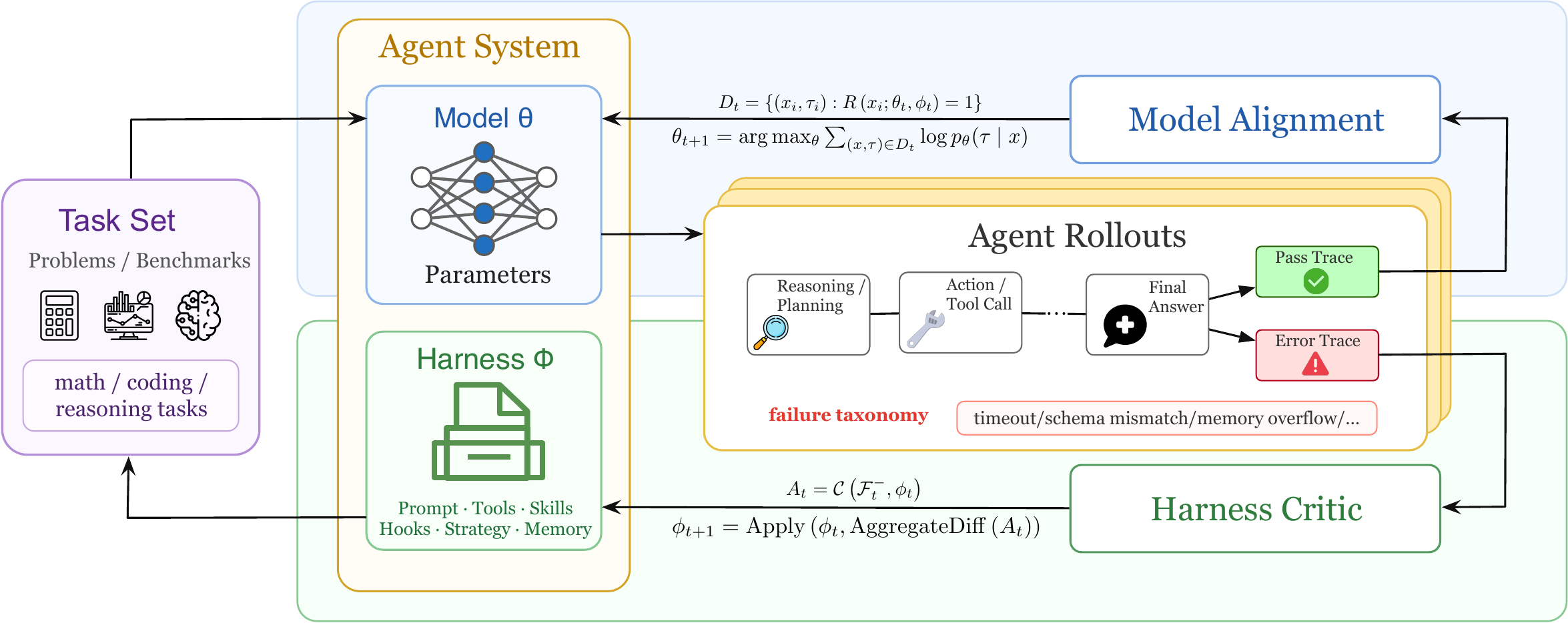}
  \caption{
    \textbf{The overall framework of Co-Harness.}
  }
  \label{fig:co-harness-framework}
\end{figure}

We propose \textbf{Co-Harness}, a simple dual-loop recipe for agent post-training.
As shown in \Cref{fig:co-harness-framework}, each round starts from a model $\thetat$ and a Harness $\phit$.
The agent first collects rollouts on a task set.
Failed trajectories are passed to an LLM-based \textbf{HarnessCritic}, which attributes each failure to a structured Harness-level cause, such as prompt ambiguity, tool schema error, missing skill, or middleware mismatch.
HarnessCritic then proposes local harness diffs, and a diff is accepted only if validation rollouts improve the targeted failure mode without regressing held-out behaviors.
The accepted harness $\phistar$ is used to collect high-quality trajectories, which are then used to fine-tune the next model $\thetavec$.
The updated model starts the next round, where it may expose new harness-level bottlenecks. The key idea is that the harness should not be treated as a fixed data-generation environment.
A better harness produces cleaner tool use, fewer avoidable crashes, and more informative trajectories; these trajectories train a stronger model, which can then exploit a richer harness:
\begin{center}
\emph{Better harness $\to$ better trajectories $\to$ stronger model $\to$ better harness $\to$ \ldots}
\end{center}
This makes the harness a learnable part of the post-training system, rather than a static scaffold around the model.
Our contributions are as follows:
\begin{enumerate}
  \item We formulate agent post-training as a joint optimization problem over the model and its harness.
  Instead of treating prompts, tools, skills, middleware, and memory as fixed infrastructure, Co-harness alternates between improving the harness from failures and training the model on trajectories generated by the improved harness.

  \item We introduce HarnessCritic, a failure-driven procedure for local harness repair.
  Given failed trajectories, HarnessCritic attributes each failure to a structured harness-level cause and proposes validated diffs across prompt, tools, skills, and middleware.

  \item We show that this simple loop gives compounding gains in tool-integrated reasoning.
  On AIME24, AIME25, and HMMT25, two rounds of Co-Harness improve average accuracy by $+20.4$ pp across Qwen3-8B and Qwen3-32B, with the largest gain of $+27.2$ pp on HMMT25 for Qwen3-32B.
  In a 200+ hour autonomous run, Co-Harness also repairs system crashes, improves inference efficiency, and discovers an ensemble strategy without human intervention.
\end{enumerate}

\section{Related Work}
\label{sec:related}

\paragraph{Tool-Integrated Reasoning (TIR).}
Tool-Integrated Reasoning augments LLMs with external computational tools---most commonly a Python code interpreter---enabling models to solve problems by interleaving text-based chain-of-thought with executable code across multiple turns~\citep{guo2025deepseek,feng2025retool,li2025torl}.
Recent work has demonstrated that TIR decisively outperforms pure-text reasoning on mathematical benchmarks, not merely by offloading arithmetic but by unlocking qualitatively new problem-solving strategies such as iterative search, DP, and hypothesis-driven exploration~\citep{yang2025tir_theory}.
Theoretically, TIR strictly expands the model's feasible support: for any finite token budget, there exist algorithmic strategies whose programmatic representation is concise while their natural-language simulation is intractably verbose~\citep{yang2025tir_theory}.
This gap makes the \emph{Harness}---the scaffolding that mediates how tool calls are formatted, executed, and fed back into reasoning---a first-order determinant of TIR performance.
Unlike pure-text agents where harness failures degrade prompt clarity, in TIR a single tool-schema error or misconfigured retry hook can abort an entire multi-turn code-interpreter session, making Harness co-evolution particularly impactful in this regime.

\paragraph{Automated Harness Design.}
More recent work expands the search space from a single prompt-like component to the full agent harness.
Meta-Harness~\citep{leeMetaHarnessEndtoEndOptimization2026} optimizes harness implementations using program source, execution traces, and task scores.
Agentic Harness Engineering~\citep{lin2026agentic} uses observability signals to revise multiple harness components, including prompts, tools, middleware, skills, and memory.
AEVO~\citep{zhang2026harnessing} frames agent evolution as an interactive process and modifies the procedure or context that steers future search.
These approaches show that harness design is a central lever for improving agent behavior without changing the base model.
In contrast, our goal is not only to find a better inference-time scaffold for a fixed model.
Co-Harness makes harness evolution part of post-training: accepted harness edits change the trajectories collected for supervision, and those trajectories are then used to update the model itself.

\paragraph{Automated design of agentic systems.}
Automated Design of Agentic Systems (ADAS) \citep{hu2024adas} proposes Meta Agent Search: a meta-agent iteratively programs new agent designs in code, using an archive of prior designs.
Agents defined in Turing-complete code can in principle represent any agentic behavior, and ADAS shows strong cross-domain transfer.
However, ADAS optimizes for \emph{in-context task accuracy} at inference time---it does not consider whether the discovered agent designs produce high-quality training trajectories, nor does it perform failure attribution.
Our work is complementary: while ADAS searches a broader design space, HarnessCritic focuses on targeted, explainable repair of known failure patterns.
A key practical difference is that HarnessCritic's search signal comes from failure trajectories (a form of guided debugging) rather than from task outcome scores alone.

\section{Method}
\label{sec:method}

Co-Harness treats Harness improvement and model improvement as two alternating processes: a better Harness surfaces cleaner trajectories that train a stronger model, which in turn exposes higher-order Harness bottlenecks previously invisible under a weaker policy.
The framework is summarized in \cref{fig:co-harness-framework}.

\subsection{Overview and Problem Formulation}
\label{sec:method-problem}

\begin{definition}[Harness configuration]
A Harness configuration $\phivec \in \Phi$ is a structured five-tuple
\[
\phivec = (\dimP, \dimT, \dimS, \dimMid, \dimM),
\]
where $\dimP$ denotes prompt and instruction templates, $\dimT$ tool definitions and interface schemas, $\dimS$ reusable skills or modules, $\dimMid$ middleware, and $\dimM$ long-term memory policy.
Here, middleware bundles the agent's runtime control surface: the orchestrator (main loop and protocol), hooks for yes/no interception and verification, and context management logic such as compression, truncation, or retrieval routing.
We write $\phit$ for the active Harness at co-evolution round $t$.
\end{definition}

Let $\thetat$ denote the model parameters at round $t$, and let $\Reward(x; \thetavec, \phivec)$ denote task reward for input $x$, instantiated in our experiments as pass@1 or resolve rate depending on the benchmark.
The ideal objective is to jointly optimize model and Harness:
\[
(\thetavec^\star, \phistar) = \argmax_{\thetavec,\,\phivec \in \Phi}
\mathbb{E}_{x \sim \mathcal{X}}\left[\Reward(x; \thetavec, \phivec)\right].
\]
Directly solving this objective is impractical because evaluating a Harness requires full agent rollouts, the reward is sparse, and changes in $\thetavec$ alter which Harness edits are even meaningful.
Our key design choice is therefore to decompose the problem into two alternating loops: one improves the Harness while holding the model fixed, and the other improves the model while holding the evolved Harness fixed.

\subsection{Dual-Loop Co-Evolution Architecture}
\label{sec:method-dual-loop}

Co-Harness alternates between \AutoHarnessLoop{} and \ModelAlignLoop{}.
At round $t$, the current model $\thetat$ interacts with the current Harness $\phit$ to produce a batch of trajectories.
Failed trajectories are routed to HarnessCritic, which proposes targeted edits to the Harness.
After validation, the accepted configuration is denoted $\phit^\star$.
Running $\thetat$ under $\phit^\star$ then yields a trajectory set $\Dt$, which becomes the training signal for the next model update.
The resulting model $\boldsymbol{\theta}_{t+1}$ is fed back into the next Co-Harness round, initialized from $\boldsymbol{\phi}_{t+1} \leftarrow \phit^\star$.

\paragraph{Co-Harness Loop.}
The Co-Harness Loop fixes $\thetat$ and improves the Harness.
Its input is a set of failures observed under the current pair $(\thetat, \phit)$.
Its output is an evolved Harness $\phit^\star$ together with a refreshed rollout set collected under that evolved configuration.
This loop is responsible for finding environment-level fixes: repairing incorrect tool schemas, introducing reusable skills, adjusting middleware behavior, or extending long-term memory when the failures are attributable to the scaffolding rather than the policy.

\paragraph{Model Alignment Loop.}
The Model Alignment Loop fixes the evolved Harness $\phit^\star$ and improves the model.
Its input is the higher-quality trajectory set $\Dt$ produced under the evolved Harness.
Its output is an updated model $\boldsymbol{\theta}_{t+1}$ that internalizes the improved behavior distribution.
This second loop is essential: without it, Harness optimization remains an inference-time search problem; with it, each successful Harness revision can become persistent model capability.

\paragraph{Alternating schedule.}
We repeat the two loops for $T$ rounds.
In practice, the schedule stops when either (i) no candidate patch passes validation, or (ii) the marginal gain in reward falls below a small threshold for consecutive rounds.
Our experiments use a fixed small number of rounds, but the formulation itself does not depend on a particular horizon.
Algorithm~\ref{alg:autoharness-round} summarizes one full round of co-evolution.

\begin{algorithm}[t]
  \caption{One round of Co-Harness co-evolution}
  \label{alg:autoharness-round}
  \begin{algorithmic}[1]
    \STATE Input current model $\thetat$ and Harness $\phit$
    \STATE Collect failed trajectories $\mathcal{F}_t^{-}$ under $(\thetat, \phit)$
    \FOR{$k = 1, \ldots, K$}
      \STATE $\mathcal{A}_t \leftarrow \Critic(\mathcal{F}_t^{-}, \phit)$
      \STATE $\DeltaPhi_t \leftarrow \textsc{AggregateDiffs}(\mathcal{A}_t)$
      \STATE $\tilde{\phi}_t \leftarrow \textsc{Apply}(\phit, \DeltaPhi_t)$
      \IF{$\textsc{Validate}(\thetat, \phit, \tilde{\phi}_t)$ is non-regressive}
        \STATE commit $\DeltaPhi_t$ to the registry and set $\phit \leftarrow \tilde{\phi}_t$
      \ELSE
        \STATE reject $\DeltaPhi_t$ and keep $\phit$
      \ENDIF
      \STATE refresh $\mathcal{F}_t^{-}$ under $(\thetat, \phit)$
    \ENDFOR
    \STATE $\Dt \leftarrow \textsc{CollectTrajectories}(\thetat, \phit)$
    \STATE $\boldsymbol{\theta}_{t+1} \leftarrow \textsc{SFT}(\thetat, \Dt)$
    \STATE \textbf{return} evolved Harness $\phit^\star \equiv \phit$, updated model $\boldsymbol{\theta}_{t+1}$
  \end{algorithmic}
\end{algorithm}

\subsection{HarnessCritic: Evolving the Harness from Failure Trajectories}
\label{sec:method-critic}

HarnessCritic, denoted $\mathcal{C}$, converts raw failed trajectories into structured evidence for Harness repair.
Given a batch $\mathcal{F}_t^{-} = \{\Trajfail_i\}$ collected under $(\thetat, \phit)$, HarnessCritic processes each trajectory and returns an \emph{attribution set} $\mathcal{A}_t$---a collection of structured records, one per failure, each specifying the predicted root cause, the implicated Harness dimension, a severity rating, supporting evidence from the trajectory, and a concrete diff suggestion.
Formally, $\mathcal{A}_t = \mathcal{C}(\mathcal{F}_t^{-}, \phit)$.
This stage lets Co-Harness optimize from recurring, explainable error patterns rather than from undirected mutation.

\paragraph{Failure attribution taxonomy.}
Table~\ref{tab:taxonomy} defines the attribution schema used throughout the paper.
The first five rows are actionable Harness categories, each tied to one of the five Harness dimensions.
HarnessCritic may also abstain with \texttt{agent\_error} when the failure is best explained by the model rather than the Harness; such cases are excluded from Harness patch generation.

\begin{table}[t]
  \centering
  \small
  \caption{Failure attribution taxonomy used by HarnessCritic.
  The first five categories are actionable Harness failures; \texttt{agent\_error} is a non-Harness abstention label.}
  \label{tab:taxonomy}
  \begin{tabular}{lll}
    \toprule
    \textbf{Root cause} & \textbf{Primary locus} & \textbf{Typical symptom} \\
    \midrule
    \texttt{prompt\_ambiguity}   & $\dimP$   & instruction underspecification or conflicting goals \\
    \texttt{tool\_schema\_error} & $\dimT$   & invalid tool call, bad argument schema, backend mismatch \\
    \texttt{skill\_missing}       & $\dimS$   & missing reusable routine or decomposition primitive \\
    \texttt{middleware\_mismatch} & $\dimMid$ & flawed loop protocol, hook behavior, or context management \\
    \texttt{memory\_overflow}     & $\dimM$   & persistent state overflow, stale memory, or retrieval failure \\
    \texttt{agent\_error}         & --         & failure remains model-side after Harness inspection \\
    \bottomrule
  \end{tabular}
\end{table}

\paragraph{Structured attribution output.}
For each failure trajectory, HarnessCritic produces a structured record of the form
\texttt{\{root\_cause, harness\_dim, severity, evidence, diff\_suggestion\}}.
The \texttt{evidence} field anchors the diagnosis to concrete trajectory events, while \texttt{diff\_suggestion} specifies a local patch target such as a tool schema field, a middleware hook insertion point, or an orchestrator policy.
Using structured outputs keeps the search space auditable and makes aggregation operate over explicit patch loci rather than over free-form natural language.

\paragraph{Aggregation into Harness Diffs.}
A single failure may be noisy, so Co-Harness aggregates attributions across a batch before editing the Harness.
We rank candidate edits by recurrence, severity, and consistency of the proposed field path, then merge compatible suggestions into a single Harness Diff $\DeltaPhi_t$.
A diff is intentionally local: it records only the fields that change, along with their old and new values.
This locality is important for both interpretability and rollback.
In contrast to open-ended code-space search, HarnessCritic performs targeted repair grounded in observed failures.

\subsection{Model Alignment Loop}
\label{sec:method-model-align}

Once the Harness has been evolved to $\phit^\star$, we re-run the current model under that configuration and collect the resulting trajectory set $\Dt$.
Each element of $\Dt$ contains the full interaction trace needed for supervision: prompt context, tool calls, tool responses, intermediate decisions, and the final verified outcome.
We keep trajectories that satisfy the task verifier or otherwise meet the quality filters associated with the benchmark.

The model update then fine-tunes $\thetat$ on $\Dt$ to obtain $\boldsymbol{\theta}_{t+1}$:
\[
\boldsymbol{\theta}_{t+1}
= \argmax_{\thetavec}
\sum_{(x,\Traj) \in \Dt} \log p_{\thetavec}(\Traj \mid x).
\]
Intuitively, this step distills the behavior enabled by the improved Harness into the model parameters.
The updated model can often solve tasks with lighter middleware intervention, shorter recovery chains, or more reliable tool use, which reduces future Harness debt.
More importantly, a stronger model exposes new failure regimes that were previously masked by weaker base behavior, giving the next Co-Harness round more meaningful problems to optimize.

\subsection{Patch Validation and Harness Registry}
\label{sec:method-validation}

A proposed Harness patch is accepted only if it improves the targeted failure mode ($\delta_{\mathrm{in}} > 0$) without causing regression on held-out behaviors ($\delta_{\mathrm{out}} \geq 0$); accepted patches are written into a versioned Harness registry that serves as an audit trail and enables full rollback (see \cref{app:harness-registry} for validation protocol and registry details).

\subsection{Why the Loop Can Compound}
\label{sec:method-compounding}

The dual loop compounds only when three conditions hold.
First, Harness updates must increase \emph{trajectory quality}, not merely change surface formatting.
Second, the model update must \emph{internalize} those improvements so that the next model is genuinely stronger rather than merely dependent on a fragile scaffold.
Third, the stronger model must \emph{unlock new Harness opportunities} by making more advanced middleware, skills, or memory mechanisms worth deploying.
If any of these conditions fails, the process collapses to a single-loop optimization with diminishing returns.

This perspective also explains why Co-Harness differs from prior inference-time Harness search.
The goal is not only to find a better external scaffold for the current model, but to create a feedback loop in which better scaffolding produces better training signal, and better models in turn enlarge the set of useful scaffolding revisions.
That mechanism is the central hypothesis tested in our experiments.

\section{Experiments}
\label{sec:experiments}

\subsection{Setup}
\label{sec:experiments-setup}

\paragraph{Task setting: Tool-Integrated Reasoning (TIR).}
All experiments adopt a TIR paradigm: the agent solves each mathematical competition problem by interleaving chain-of-thought reasoning with iterative calls to a Python code interpreter.
Harness deficiencies---such as an incorrect tool schema or a missing retry hook---translate directly into failed trajectories that the model cannot recover from through reasoning alone, making TIR a particularly sharp testbed for Harness co-evolution (see \cref{app:tir-setup} for the full loop structure and failure mode taxonomy).

\paragraph{Benchmarks.}
We evaluate on three competitive mathematical reasoning benchmarks of varying difficulty:
AIME 2024 (30 problems), AIME 2025 (30 problems), and HMMT February 2025 (30 problems from the individual round).
These benchmarks span a range from moderately challenging (AIME24) to extremely difficult (HMMT25), allowing us to assess how Harness evolution benefits vary with task difficulty.
All problems are solved under the TIR setting described above; the reported accuracy is pass@1, averaging over multiple rollouts per problem.

\paragraph{Base models.}
We experiment with two model scales:
Qwen3-8B and Qwen3-32B.
Both models have strong mathematical reasoning priors and support tool-augmented generation, making them representative of practical TIR agent deployment for reasoning tasks.

\paragraph{Co-evolution rounds.}
For the dual-loop experiment, we run $T=2$ full co-evolution rounds (Round 1 and Round 2), each consisting of one Co-Harness Loop pass (HarnessCritic with $K=5$ inner iterations) followed by one Model Alignment Loop pass (SFT on the collected trajectory set).
Round 0 (Baseline) applies one HarnessCritic pass to obtain $\phivec_0^*$ but performs no SFT, and serves as the starting point for measuring compounding gains.

\paragraph{Baselines.}
\begin{itemize}
  \item \textbf{Baseline (Round 0)}: Harness evolved by one HarnessCritic pass ($\phivec_0^*$), but no SFT model alignment.
  \item \textbf{Human}: Human-designed static Harness with no model alignment, serving as an upper bound for manual engineering.
  \item \textbf{Co-Harness (ours)}: Full dual-loop co-evolution as described in \cref{sec:method-dual-loop}.
\end{itemize}

\subsection{Dual-Loop Co-Evolution: Core Experiment}
\label{sec:experiments-dual-loop}

\begin{figure}[t]
\centering
\includegraphics[width=\linewidth]{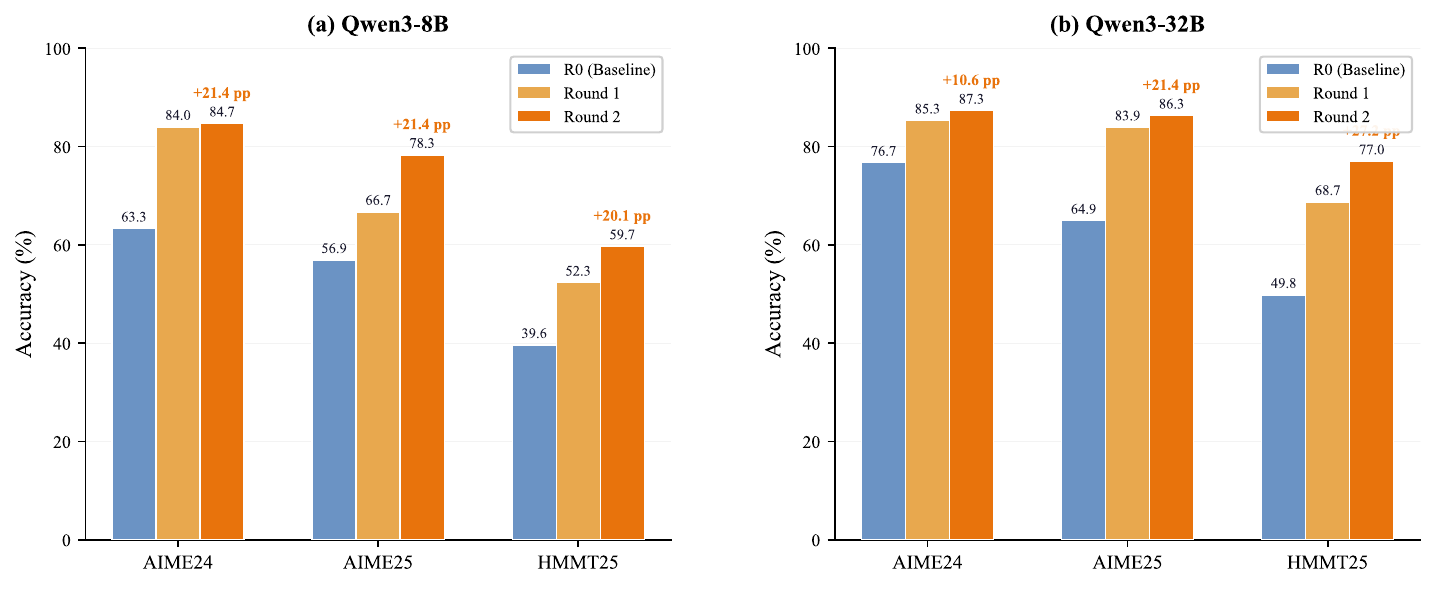}
\caption{
  \textbf{Co-Harness dual-loop co-evolution results across benchmarks and model scales.}
  Grouped bar charts showing accuracy (\%) on AIME24, AIME25, and HMMT25 for Qwen3-8B (left) and Qwen3-32B (right) across three co-evolution rounds.
  Each group contains three bars: Baseline (Round 0, Harness evolved, no SFT), Round 1 (one full co-evolution round: HarnessCritic + SFT), and Round 2 (two full rounds).
  Delta values above Round 2 bars indicate cumulative improvement over Baseline.
  Co-Harness consistently improves accuracy across all benchmarks and model scales, with larger gains on harder datasets (HMMT25).
}
\label{fig:main-results}
\end{figure}

\begin{table}[t]
\centering
\caption{
  \textbf{Co-Harness dual-loop co-evolution results.}
  Accuracy (\%) on mathematical reasoning benchmarks.
  \textbf{Human}$^\dagger$: Human-designed static Harness (no model alignment).
  \textbf{R0}: Baseline --- HarnessCritic-evolved Harness ($\phivec_0^*$), no SFT model alignment.
  \textbf{R1}: After one round of HarnessCritic evolution + SFT.
  \textbf{R2}: After two rounds (ours).
  $\boldsymbol{\Delta}$\textbf{(R0)}: cumulative gain of R2 over Baseline.
  $\boldsymbol{\Delta}$\textbf{(Hu)}: gain of R2 over Human-designed Harness.
  Best results in \textbf{bold}.
}
\label{tab:dual-loop}
\small
\begin{tabular}{llcccccc}
\toprule
\textbf{Model} & \textbf{Benchmark} & \textbf{Human}$^\dagger$ & \textbf{R0} & \textbf{R1} & \textbf{R2} & $\boldsymbol{\Delta}$\textbf{(R0)} & $\boldsymbol{\Delta}$\textbf{(Hu)} \\
\midrule
\multirow{3}{*}{Qwen3-8B}
  & AIME24        & 59.3 & 63.3 & 84.0 & \textbf{84.7} & +21.4 & +25.4 \\
  & {AIME25} & 51.3 & 56.9 & 66.7 & \textbf{78.3} & +21.4 & +27.0 \\
  & HMMT25        & 34.7 & 39.6 & 52.3 & \textbf{59.7} & +20.1 & +25.0 \\
\midrule
\multirow{3}{*}{Qwen3-32B}
  & AIME24        & 72.0 & 76.7 & 85.3 & \textbf{87.3} & +10.6 & +15.3 \\
  & {AIME25} & 61.3 & 64.9 & 83.9 & \textbf{86.3} & +21.4 & +25.0 \\
  & HMMT25        & 46.7 & 49.8 & 68.7 & \textbf{77.0} & +27.2 & +30.3 \\
\midrule
\multicolumn{2}{l}{\textbf{Average}} & \textbf{54.2} & \textbf{58.5} & \textbf{73.5} & \textbf{78.9} & \textbf{+20.4} & \textbf{+24.7} \\
\bottomrule
\end{tabular}
\end{table}

\cref{fig:main-results} and \cref{tab:dual-loop} present the core experiment.
We observe three key findings:

\paragraph{Consistent compounding gains and surpassing human-designed Harness.}
Co-Harness produces monotonic accuracy improvements across rounds on both model scales and all three benchmarks.
The average accuracy improves from 58.5\% (Baseline) to 73.5\% (Round 1, +15.0 pp) to 78.9\% (Round 2, +20.4 pp cumulative), confirming that the dual-loop co-evolution mechanism generates genuine compounding self-improvement.
Crucially, R2 also surpasses the Human-designed static Harness by an average of +24.7 pp---demonstrating that the automated co-evolution not only improves over its own starting point but also exceeds the ceiling of a carefully hand-crafted, fixed configuration.

\paragraph{Harder benchmarks benefit more.}
The gains from Harness co-evolution are largest on HMMT25 (the most difficult benchmark): +20.1 pp for 8B and +27.2 pp for 32B.
This confirms the core insight that better Harness configurations are most impactful when the model faces complex, multi-step reasoning tasks where scaffolding deficiencies are more likely to cause failures.
In contrast, AIME24 (where baselines are already relatively high) shows the smallest gains for both models.

\paragraph{Larger models benefit from Harness co-evolution more on hard tasks.}
For the 32B model, HMMT25 improvement (+27.2 pp) substantially exceeds AIME24 improvement (+10.6 pp).
This indicates that a stronger model has greater latent capacity that can be unlocked by better Harness---precisely the mechanism described in \cref{sec:method-compounding}.

\subsection{Autonomous Harness Evolution: Case Study on AIME24}
\label{sec:experiments-case-study}

\begin{figure}[t]
\centering
\includegraphics[width=\linewidth]{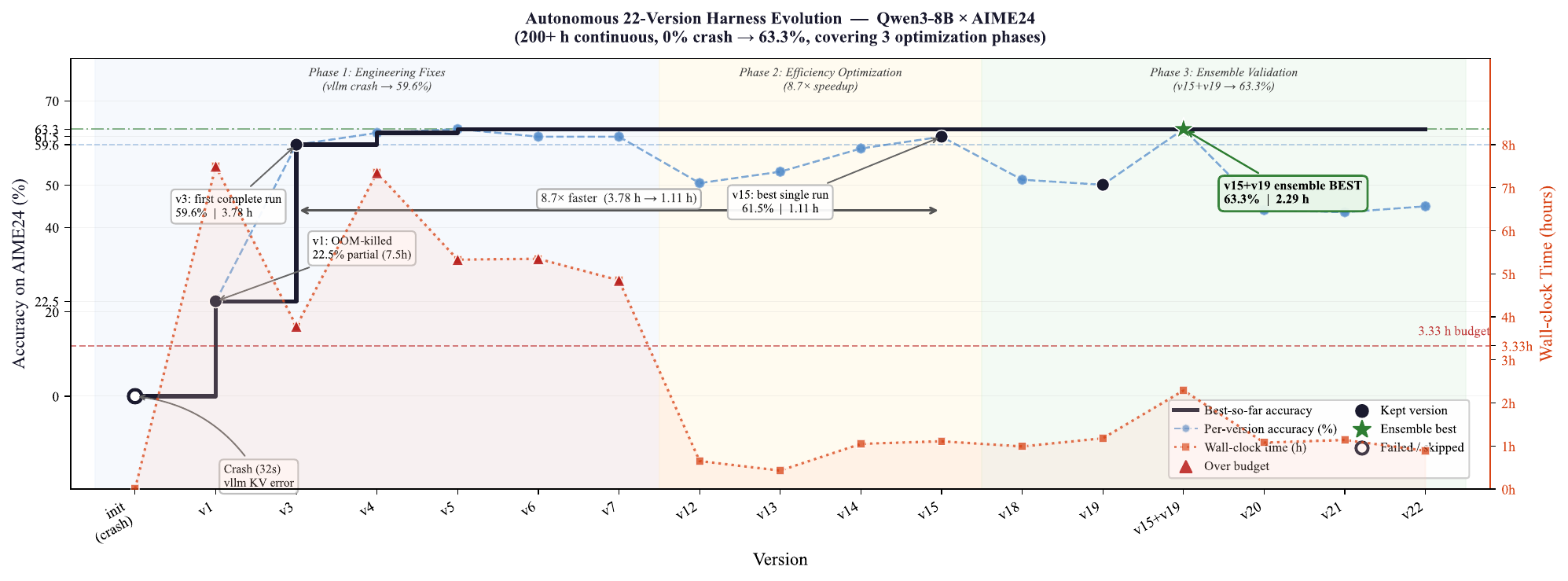}
\caption{
  \textbf{Autonomous 22-version Harness evolution on AIME24 with Qwen3-8B.}
  The agent ran continuously for over 200 hours without human intervention, progressing through three optimization phases:
  \textit{Phase~1} (engineering fixes, 0\% $\to$ 59.6\%),
  \textit{Phase~2} (efficiency optimization, $8.7\times$ speedup),
  and \textit{Phase~3} (ensemble validation, 63.3\% final accuracy).
  Left axis: accuracy (\%); right axis: wall-clock time (hours).
  Red dashed line: 3.33\,h time budget.
  Green star: best ensemble result (v15+v19).
}
\label{fig:aime-evolution}
\end{figure}

\cref{fig:aime-evolution} shows a complete, autonomous 22-version Harness evolution trajectory on AIME24 with Qwen3-8B.
Starting from a crashed initial configuration (vllm KV cache error in 32 seconds), the Co-Harness Loop autonomously progressed through three phases:

\begin{enumerate}
  \item \textbf{Phase~1 --- Engineering fixes (init--v7)}: HarnessCritic attributed the initial crash to a \texttt{tool\_schema\_error}, patched the launch configuration, and resolved a zombie thread bug by switching from \texttt{ThreadPool} to \texttt{ProcessPool}. The first complete evaluation (v3) achieved 59.6\% at 3.78\,h.
  \item \textbf{Phase~2 --- Efficiency optimization (v12--v15)}: Global-batched inference reduced wall-clock time by $8.7\times$ (3.78\,h $\to$ 1.11\,h), while restoring chain-of-thought on follow-up turns maintained accuracy at 61.5\%.
  \item \textbf{Phase~3 --- Ensemble validation (v18--v22)}: Combining two diverse seeds via 6-trajectory majority voting achieved the best result of \textbf{63.3\%} at 2.29\,h total---the highest accuracy while remaining within the 3.33\,h time budget.
\end{enumerate}

This trajectory demonstrates that HarnessCritic can autonomously navigate the accuracy--efficiency trade-off, detect and roll back regressions (v20--v22: domain-specific prompts that conflicted with Qwen3's internal reasoning), and discover non-obvious improvements such as ensemble strategies.

\section{Analysis and Discussion}
\label{sec:analysis}

\paragraph{8B saturation effects.}
For Qwen3-8B on AIME24, Round 2 (84.7\%) shows negligible improvement over Round 1 (84.0\%).
We attribute this to two factors:
(1) the 8B model's reasoning capacity approaches its ceiling on this benchmark, limiting the benefit of further Harness evolution;
(2) the high R0 baseline (63.3\%) means fewer remaining Harness-attributable failures---most errors are now model-internal rather than scaffolding-related.
In contrast, the 32B model continues improving on AIME24 (R1 $\to$ R2: +2.0 pp), as its stronger reasoning can better exploit the evolved Harness.

\paragraph{When does co-evolution fail to compound?}
Three failure modes are empirically observed:
\textit{(1) SFT plateau}: if the model's architecture limits further reasoning improvement, SFT gains saturate and the next Co-Harness round re-enters the same failure distribution.
We observe hints of this for 8B on AIME24.
\textit{(2) Harness over-patching}: excessive Diff acceptance in early rounds can introduce complexity that overwhelms a weaker model.
\textit{(3) Attribution cascade}: a Diff that repairs one deficiency while introducing another creates a more complex failure distribution for the next round; the version-controlled Harness registry allows rollback to the last validated configuration.

\paragraph{Harness Debt analysis.}
A key concern for trajectory-based training is \emph{Harness Debt}: if trajectories are collected under a scaffolding configuration that compensates for model weaknesses, the trained model may over-rely on that scaffolding and under-perform at test time (without Harness).
Our results show that Co-Harness \emph{reduces} Harness Debt: the model trained on HarnessCritic-evolved trajectories achieves higher absolute accuracy at test time each round, demonstrating that the trajectories build transferable reasoning skills rather than Harness-dependent behaviors.
Notably, the 8B model's AIME25 accuracy improves from 56.9\% (Baseline) to 78.3\% (Round 2)---a +21.4 pp gain that reflects genuine skill acquisition rather than mere Harness memorization.

\section{Conclusion}
\label{sec:conclusion}

We introduced \textbf{Co-Harness}, a dual-loop co-evolution framework for language agents.
The framework interleaves two loops: the \emph{Co-Harness Loop}, which uses HarnessCritic to evolve the agent scaffolding via semantically-grounded failure attribution, and the \emph{Model Alignment Loop}, which fine-tunes the model on the high-quality trajectories produced by the evolved Harness.
The two loops create a positive feedback spiral---better Harness produces better trajectories, which train a stronger model, which in turn unlocks further Harness improvements previously inaccessible.

On Tool-Integrated Reasoning (TIR) benchmarks (AIME24, AIME25, HMMT25), two rounds of Co-Harness dual-loop co-evolution yield an average accuracy improvement of $+20.4$ pp across Qwen3-8B and Qwen3-32B, with the largest gain of $+27.2$ pp on HMMT25 (the hardest benchmark) for Qwen3-32B---confirming that better Harness produces better trajectories, which train stronger models, which unlock further Harness improvements.

The core insight---that the scaffolding and the model are not independent but interdependent axes of optimization---opens a new direction for agent training.
Current paradigms either freeze the Harness (SWE-smith, AgentTuning) or freeze the model (ADAS, OPRO); Co-Harness is the first framework to co-evolve both, realizing the compounding self-improvement that neither alone can achieve.

\paragraph{Limitations.}
Co-Harness requires a capable critic LLM, a minimum volume of failure trajectories for cold-start, and significant compute for multi-round SFT.
Attribution accuracy degrades for failure modes requiring counterfactual control-flow reasoning.
Structural Harness redesign remains a human responsibility.

\paragraph{Future directions.}
Three extensions are particularly promising.
\emph{Online Co-Harness} would perform HarnessCritic attribution in real time during rollout, enabling within-run Harness adaptation for long-horizon tasks.
\emph{RL-based Harness evolution} would replace the LLM critic with a reward-signal-driven search over the Harness space, potentially discovering non-obvious configurations.
\emph{Multi-agent Co-Harness} would extend co-evolution to systems with multiple specialized agents, where the inter-agent middleware layer (the $\dimMid$ dimension) becomes a key optimization target.
Together, these extensions would move Co-Harness from a periodic batch process toward a continuous, online scaffolding health monitor that evolves alongside the model.

\bibliographystyle{unsrtnat}
\bibliography{references}

\newpage
\appendix

\section{TIR Agent Setup: Multi-Turn Code Interpreter Loop}
\label{app:tir-setup}

All experiments in this paper use a Tool-Integrated Reasoning (TIR) paradigm, where the agent solves mathematical competition problems by interleaving chain-of-thought reasoning with iterative Python code interpreter calls.
We describe the TIR loop structure and how the Harness mediates it.

\paragraph{TIR loop structure.}
Each episode begins with the problem statement inserted into the system prompt.
The model then generates a response that may include zero or more \emph{tool call} blocks of the form:
\begin{verbatim}
<tool_call>
{"name": "python", "arguments": {"code": "..."}}
</tool_call>
\end{verbatim}
The Middleware intercepts each tool call, dispatches the code block to the Python interpreter subprocess, and appends the interpreter's response (stdout, stderr, and return value) back into the conversation context as a \emph{tool result}:
\begin{verbatim}
<tool_result>
{"stdout": "42\n", "stderr": "", "exit_code": 0}
</tool_result>
\end{verbatim}
The model then continues generating, optionally issuing further tool calls or producing its final answer.
This process repeats for up to \texttt{max\_turns} turns (initially 15, increased to 30 after Phase~2 harness evolution in the case study).

\paragraph{How TIR trajectories reveal Harness deficiencies.}
Failed TIR trajectories exhibit failure modes that are qualitatively different from pure-text failures:
\begin{itemize}
  \item \textbf{Tool-schema failures ($\dimT$)}: the tool call JSON is malformed or uses an unsupported argument name, causing the interpreter to reject the call with a schema validation error rather than executing the code. This produces an unhelpful error token in the context and often stalls the episode.
  \item \textbf{Middleware failures ($\dimMid$)}: the retry hook is absent or misconfigured, so a transient interpreter crash (e.g., subprocess OOM) causes the episode to terminate silently rather than triggering a retry. The model receives no feedback signal.
  \item \textbf{Context management failures ($\dimMid$)}: in long episodes where the model issues many tool calls, accumulated interpreter outputs overflow the context window. A poorly configured compression policy discards critical intermediate results, causing the model to repeat calculations or lose track of partial progress.
  \item \textbf{Prompt failures ($\dimP$)}: the system prompt does not clearly instruct the model to use code as a \emph{primary} reasoning tool, leading the model to defer code calls until the final step, missing the opportunity for iterative exploration and verification.
\end{itemize}
HarnessCritic attributes each failed trajectory to one of these categories and generates a targeted patch.
The AIME24 case study (\cref{app:case-study-aime}) illustrates all four failure modes in a single autonomous evolution run.

\paragraph{TIR harness vs.\ pure-text harness.}
In a pure-text agent, a suboptimal prompt degrades response quality gradually; the model can often produce a usable trajectory despite imperfect instructions.
In TIR, a single misconfigured component can abort the episode entirely: if the Python interpreter rejects the tool call, the model receives no computational result and is forced to guess or produce a wrong answer.
This binary nature of TIR failures---either the code executes and returns useful feedback, or it does not---creates a sharper gradient for HarnessCritic to act on, and explains why Harness co-evolution yields particularly large gains in this regime.

\section{Attribution Prompt Template}
\label{app:prompt}

The following prompt template is used for LLM-based failure attribution within HarnessCritic.
Chain-of-thought reasoning (\texttt{<think>} block) precedes the structured JSON output.

\begin{small}
\begin{verbatim}
System: You are an expert agent scaffolding engineer.
Given a failed agent trajectory and the current Harness
configuration, identify the root cause of failure.

Output your reasoning in <think>...</think>, then output
a JSON object matching this schema:
{
  "root_cause": one of [
    "prompt_ambiguity", "tool_schema_error", "skill_missing",
    "middleware_mismatch", "memory_overflow", "agent_error"
  ],
  "harness_dim": one of ["P", "T", "S", "Mid", "M"],
  "severity": one of ["critical", "major", "minor"],
  "evidence": "< 1 sentence citing specific trajectory event >",
  "diff_suggestion": {
    "field_path": "middleware.hooks.post_tool_call[0].type",
    "old_value": null,
    "new_value": "retry_on_rate_limit"
  }
}

== CURRENT HARNESS CONFIGURATION ==
{harness_config_yaml}

== FAILED TRAJECTORY ==
{trajectory_text}
\end{verbatim}
\end{small}

\section{Harness Configuration Details}
\label{app:harness-config}

The Full-Harness initial configuration $\phivec_0$ used in all experiments.
Note that the Tool dimension is configured for TIR: the primary tool is a Python code interpreter, and the Middleware is responsible for managing the multi-turn code-execution loop.

\begin{table}[h]
\centering
\small
\caption{Initial Harness $\phivec_0$ configuration (TIR setting).}
\begin{tabular}{lp{8cm}}
\toprule
\textbf{Dimension} & \textbf{Setting} \\
\midrule
\dimP (Prompt) & System prompt: 512-token instruction with mathematical reasoning guidelines and TIR usage instructions (``use Python code to assist your reasoning''); no few-shot retrieval \\
\dimT (Tool) & Python code interpreter (sandboxed subprocess); JSON tool-call schema specifying \texttt{name: python}, \texttt{arguments.code: str}; default timeout 30s \\
\dimS (Skill) & None (empty skill set at Round 0) \\
\dimMid (Middleware) & Orchestrator: max turns; terminate on \texttt{<DONE>} or final-answer token. Hooks: retry on ToolCallException (max 3, linear backoff). Context management: sliding-window compression keeping the last 4k tokens; interpreter outputs truncated to 2k tokens \\
\dimM (Memory) & No long-term memory or external retrieval store at Round 0 \\
\bottomrule
\end{tabular}
\end{table}

\section{Human Annotation Protocol}
\label{app:annotation}

Two expert annotators (with $>$2 years of agent engineering experience) independently labeled 200 failure trajectories using the taxonomy in \cref{tab:taxonomy}.
Disagreements were resolved by a third expert.
Human inter-annotator agreement was $\kappa = 0.77$.
Annotators were given 90 minutes for the full 200 trajectories, with access to the Harness configuration and full trajectory logs.
No information about LLM attributions was shown to annotators.
All attribution experiments use the base model $\thetavec_0$ (Round 0).

\section{Patch Validation Protocol and Harness Registry}
\label{app:harness-registry}

\paragraph{Validation protocol.}
A proposed Harness patch $\tilde{\phi}_t$ is validated against the current Harness $\phit$ on two held-out splits: a held-in set $\mathcal{V}_{\mathrm{in}}$ containing examples that exhibit the target failure mode, and a held-out set $\mathcal{V}_{\mathrm{out}}$ containing previously successful or orthogonal behaviors.
We compute
\[
\delta_{\mathrm{in}} = \Reward(\mathcal{V}_{\mathrm{in}}; \thetat, \tilde{\phi}_t) - \Reward(\mathcal{V}_{\mathrm{in}}; \thetat, \phit),
\]
\[
\delta_{\mathrm{out}} = \Reward(\mathcal{V}_{\mathrm{out}}; \thetat, \tilde{\phi}_t) - \Reward(\mathcal{V}_{\mathrm{out}}; \thetat, \phit).
\]
A patch is accepted only if $\delta_{\mathrm{in}} > 0$ and $\delta_{\mathrm{out}} \geq 0$.
This rule prevents the system from overfitting to isolated failures or trading away previously reliable behavior.
Rejected patches are rolled back immediately; the registry audit trail records both accepted and rejected candidates for reproducibility.

\paragraph{Harness registry.}
All Harness configurations across co-evolution rounds are stored in a versioned Harness registry using a simple directory structure:

\begin{verbatim}
harness_registry/
  v0/  harness.yaml          # phi_0 (initial)
  v1/  harness.yaml          # phi_0^* (after Round 1 HarnessCritic)
       diffs/
         001_hook_retry.yaml
         002_tool_timeout.yaml
         003_prompt_scope.yaml
         004_strategy_maxrounds.yaml
  v2/  harness.yaml          # phi_1^* (after Round 2 HarnessCritic)
       diffs/ ...
  v3/  harness.yaml          # phi_2^* (after Round 3 HarnessCritic)
       diffs/ ...
\end{verbatim}

Each diff file is a self-contained YAML patch with metadata (round, attribution source, validation delta) that enables full rollback and experiment reproducibility.
The registry also serves as an interpretable audit trail of what changed across co-evolution rounds and why.

\section{Extended Analysis}
\label{app:extended-analysis}

\paragraph{Why does Harness co-evolution matter specifically in TIR?}
The TIR paradigm makes the Harness a \emph{structural bottleneck}: the Harness mediates every step of the code-interpreter loop, from how tool calls are serialized and dispatched to how execution results are routed back into the model's context and how errors trigger retries.
A pure-text reasoning agent can fall back to verbal computation if a prompt is suboptimal; a TIR agent that encounters an invalid tool schema or a missing retry hook simply fails---the interpreter rejects the call, no result is returned, and the episode terminates with no signal for the model to learn from.
HarnessCritic's attribution-guided repair targets precisely these components---tool schemas, middleware hooks, turn limits---whose deficiencies cause TIR trajectories to fail regardless of model capability, explaining why even minor tool-interface fixes can unlock entire classes of previously unreachable multi-turn strategies.

\paragraph{Which benchmarks benefit most from Harness evolution?}
The largest accuracy improvements occur on the hardest benchmark (HMMT25: +20.1 pp for 8B, +27.2 pp for 32B).
Complex, multi-step TIR problems expose more Harness deficiencies---inadequate retry policies for long code-interpreter sessions, context overflow when multiple execution results accumulate, and strategy mismatches in middleware routing---and harder problems also require more code-interpreter turns per episode, amplifying the cost of any single harness deficiency.
Simpler benchmarks (AIME24) show smaller but still meaningful gains, as fewer Harness-attributable failure modes remain and the model can often compensate with fewer code calls.

\paragraph{Model scale and co-evolution potential.}
The 32B model shows \emph{larger} absolute gains on hard benchmarks (HMMT25: +27.2 pp) than the 8B model (+20.1 pp), indicating that a stronger model has greater latent capacity that can be unlocked by better Harness---the compounding mechanism described in \cref{sec:method-compounding}.
Both models achieve comparable gains on AIME25 (+21.4 pp each), suggesting that the Harness--model interaction depends on the specific failure distribution at each benchmark--model combination.

\section{Co-Evolution Trajectory Analysis}
\label{app:trajectory-analysis}

\begin{figure}[h]
\centering
\includegraphics[width=\linewidth]{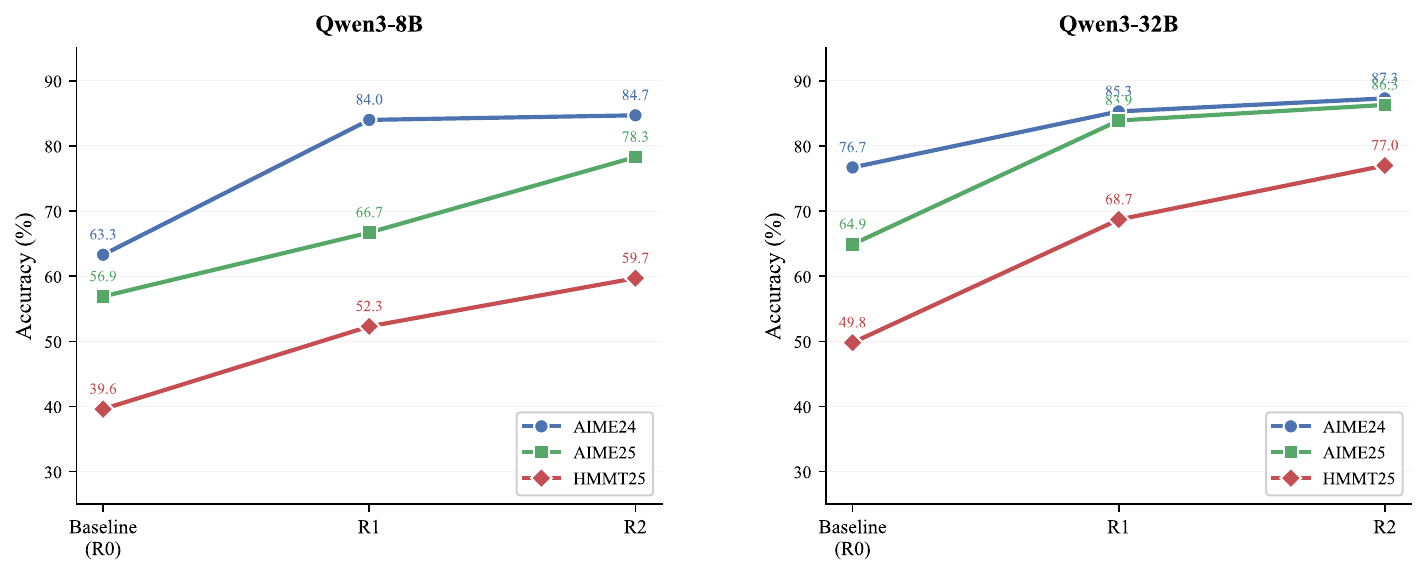}
\caption{
  \textbf{Accuracy vs.\ co-evolution round.}
  Line plots showing accuracy progression across rounds for each benchmark.
  Qwen3-8B (left) and Qwen3-32B (right).
  All curves show consistent upward trends, with the steepest gains typically occurring between Baseline and Round 1.
}
\label{fig:rounds-trend}
\end{figure}

\cref{fig:rounds-trend} visualizes the per-benchmark accuracy trajectory across co-evolution rounds.
Two patterns emerge:

\paragraph{Diminishing but persistent gains.}
The largest accuracy jump occurs between Baseline and Round 1 (average +15.0 pp), with a smaller but meaningful gain from Round 1 to Round 2 (+5.4 pp).
This is consistent with the compounding condition analysis: the easiest Harness deficiencies are fixed first, and subsequent rounds address increasingly subtle failure modes.

\paragraph{8B saturation on AIME24.}
For Qwen3-8B on AIME24, Round 2 (84.7\%) shows minimal improvement over Round 1 (84.0\%), suggesting that the 8B model approaches its reasoning ceiling on this benchmark.
In contrast, the 32B model continues to improve (+2.0 pp from R1 to R2 on AIME24), indicating that a stronger model can better exploit the evolved Harness.

\section{Case Study: Autonomous 22-Version Evolution on AIME24}
\label{app:case-study-aime}

To illustrate how the Co-Harness Loop behaves in a single extended run, we trace a complete evolution trajectory on the AIME 2024 mathematical reasoning benchmark with Qwen3-8B.
The agent ran autonomously for over 200 hours, producing 22 Harness versions without human intervention.
The full trajectory is shown in \cref{fig:aime-evolution}; we provide a detailed per-phase analysis here.

\paragraph{Phase~1: Engineering fixes (versions init--v7).}
The initial configuration (\texttt{init}) crashed in 32 seconds due to a \texttt{gpu\_util} parameter mismatch in the vllm backend that caused the KV cache to be allocated below the minimum threshold.
HarnessCritic attributed this to a \texttt{tool\_schema\_error} and patched the launch configuration.
The next run (v1) exposed a zombie thread bug that caused OOM at question 55 after 7.5 hours (partial score: 22.5\%).
After switching from \texttt{ThreadPool} to \texttt{ProcessPool} with \texttt{SIGKILL}-based cleanup and enabling chain-of-thought (\texttt{thinking=ON}), v3 completed all 90 questions at 59.6\% in 3.78 hours---the first valid evaluation.
Subsequent variants (v4--v7) explored higher sampling counts ($n\!=\!5$), system prompt variants, and temperature schedules, reaching up to 63.3\% accuracy (v5) but all exceeding the 3.33-hour time budget.

\paragraph{Phase~2: Efficiency optimisation (versions v12--v15).}
HarnessCritic identified the sequential per-question execution as the primary latency bottleneck.
Switching to global-batched inference (v12--v14) reduced wall-clock time from 3.78\,h to under 1.1\,h---an $8.7\times$ speedup---at the cost of a transient accuracy drop (v12: 50.5\%, attributed to inadvertently disabling chain-of-thought on follow-up turns).
Restoring \texttt{thinking=ON} throughout and extending \texttt{max\_turns} from 15 to 30 recovered accuracy (v15: 61.5\%, 1.11\,h), now comfortably within budget.

\paragraph{Phase~3: Ensemble validation (versions v18--v22).}
With a fast and budget-compliant single run established, HarnessCritic shifted focus to trajectory diversity.
Running an independent seed (v19: \texttt{seed=42}, 1.18\,h, solo accuracy 50.1\%) and combining it with v15 via 6-trajectory majority voting raised the ensemble score to \textbf{63.3\%} at 2.29\,h total---the highest accuracy achieved while remaining within budget.
Further attempts to add a domain-specific system prompt (v20--v22) degraded accuracy by $\sim$16.5~pp, as the injected instructions conflicted with Qwen3's internal reasoning format; HarnessCritic correctly identified this as a \texttt{prompt\_ambiguity} regression and rolled back.

\paragraph{Takeaway.}
This trajectory demonstrates three properties of the Co-Harness Loop in a naturalistic setting:
(1) \emph{correctness-first repair}---the loop reliably fixes structural crashes and partial failures before optimising performance metrics;
(2) \emph{Pareto-aware search}---HarnessCritic navigates the accuracy--efficiency trade-off by treating wall-clock time as an explicit constraint, not just a metric;
(3) \emph{rollback safety}---regressions from over-patching (v20--v22) are detected by validation re-rollout and reverted, preventing the final configuration from degrading.

\end{document}